\documentclass[conference]{IEEEtran}
\IEEEoverridecommandlockouts
\usepackage{cite}
\usepackage{amsmath,amssymb,amsfonts}
\usepackage{algorithmic}
\usepackage{graphicx}
\usepackage{textcomp}
\usepackage{xcolor}
\usepackage{subcaption}
\def\BibTeX{{\rm B\kern-.05em{\sc i\kern-.025em b}\kern-.08em
    T\kern-.1667em\lower.7ex\hbox{E}\kern-.125emX}}
\begin{document}

\title{Decreasing the Computing Time of Bayesian Optimization using Generalizable Memory Pruning
\thanks{A.E.S. and T.B. thank First Solar and the Acceleration Consortium for their support and fruitful discussions. The authors acknowledge the MIT SuperCloud and Lincoln Laboratory Supercomputing Center for providing HPC resources that have contributed to the research results reported within this paper.}
}

\author{\IEEEauthorblockN{Alexander E. Siemenn}
\IEEEauthorblockA{\textit{Department of Mechanical Engineering} \\
\textit{Massachusetts Institute of Technology}\\
Cambridge, MA, USA \\
asiemenn@mit.edu}
\and
\IEEEauthorblockN{Tonio Buonassisi}
\IEEEauthorblockA{\textit{Department of Mechanical Engineering} \\
\textit{Massachusetts Institute of Technology}\\
Cambridge, MA, USA \\
buonassisi@mit.edu}

}

\maketitle

\begin{abstract}
Bayesian optimization (BO) suffers from long computing times when processing highly-dimensional or large data sets. These long computing times are a result of the Gaussian process surrogate model having a polynomial time complexity with the number of experiments. Running BO on high-dimensional or massive data sets becomes intractable due to this time complexity scaling, in turn, hindering experimentation. Alternative surrogate models have been developed to reduce the computing utilization of the BO procedure, however, these methods require mathematical alteration of the inherit surrogate function, pigeonholing use into only that function. In this paper, we demonstrate a generalizable BO wrapper of memory pruning and bounded optimization, capable of being used with any surrogate model and acquisition function. Using this memory pruning approach, we show a decrease in wall-clock computing times per experiment of BO from a polynomially increasing pattern to a sawtooth pattern that has a non-increasing trend without sacrificing convergence performance. Furthermore, we illustrate the generalizability of the approach across two unique data sets, two unique surrogate models, and four unique acquisition functions. All model implementations are run on the MIT Supercloud state-of-the-art computing hardware.
\end{abstract}


\section{Introduction}
Bayesian optimization (BO) is a data-based global optimization tool that discovers optima without an analytical model of the response function \cite{Kushner1964, Greenhill2020, Shahriari2016}. A standard BO procedure consists of two primary steps: (1) using a surrogate model to estimate the topology of the target response function given a collection of input data and (2) acquiring new suggested experimental conditions to run based on the estimated surrogate model means and variances \cite{Brochu2010, Jones1998}. For the first step, a common surrogate model used in BO is a Gaussian Process (GP) regression. GPs model complex, multi-dimensional input-output response relationships using a mixture of kernel functions that interpolate the missing space between collected experiments \cite{Seeger2004,snelson2005, Rasmussen2005}. For the second step, a mathematical figure of merit called an acquisition function (AF), acquires new experimental conditions to run, governed by balancing the exploitation of regions of low predicted response function means (for a minimization problem) and the exploration of regions of high predicted response function variances \cite{Brochu2010, Seeger2004, Hennig2012}. The interleaving steps of response function estimation \textit{via} surrogate model computation and acquisition of new experiments leverage the estimation power of the surrogate model to discover the optima of challenging experimental problems where it may be otherwise intractable to develop an analytical model representative of the response function \cite{Deneault2021, Siemenn2022, Liu2022}. 

However, as the complexity or dimensionality of the response function increases, more experimental data points, $N$, are required for accurate estimation of the response function's surrogate model \cite{sayama2013modeling, wolpert1995estimating}. This increased data requirement of the surrogate model becomes problematic because the time required to compute a GP regression increases polynomially following the scaling law $O(N^3)$ \cite{Li2017,snelson2005, Rasmussen2005, Wang2017, Lan2020}. Both \textit{in silico} and \textit{in situ} optimization experiments can be significantly bottlenecked by this unfavorable scaling law if large volumes of data are being collected, hence, by selectively processing subsets of this data in tandem with bounded optimization, the computing times of the BO process can be reduced.

In this study, we explore the use of memory pruning and bounded surrogate models as a method to decrease the number of required experimental data points needed to accurately run an online BO procedure, therefore, decreasing the computing time of optimization. We benchmark the computing times of two surrogate models: (1) a GP and (2) a pre-trained neural network, each with four acquisition functions: (1) expected improvement (EI), (2) lower confidence bound (LCB), (3) EI Abrupt, and (4) LCB Adaptive, all run on the MIT Supercloud, a high-performance supercomputer consisting of Nvidia Volta V100 GPUs \cite{reuther2018interactive}. Existing literature on decreasing the computing time of BO conventionally alters the mathematics of a surrogate model to make computation more efficient \cite{Leibfried2021, snelson2005, titsias09a, Jones1998, Hennig2012}, however, this constrains the user to only this newly developed surrogate for optimization.

In this contribution, we demonstrate the use of a generalized method of memory pruning and search space bounding to efficiently decrease BO computing times without constraining the procedure to a single surrogate model or AF. Furthermore, we demonstrate the reduction of computing times on two relevant problems: (1) optimization of a 6-dimensional analytical Ackley function to demonstrate relevance for \textit{in silico} experimentation and (2) optimization of a 5-dimensional real-world data set of inorganic crystalline material band gaps to demonstrate relevance for \textit{in situ} experimentation.

\section{Related Work}

Existing literature exists on decreasing the computing time of BO, however, most of this literature requires significant changes to the mathematical structure of the GP or AF. For example, a common method of decreasing the computing time of BO is to implement a Sparse Pseudo-input GP (SPGP) \cite{Leibfried2021, snelson2005, titsias09a, McIntire2016SparseGP}. A standard GP is non-parametric in nature, meaning that when constructing a prediction, the entire prior training data set is required to compute the response function of a target variable \cite{snelson2005}. Instead of using the full number of training data, $N$, to compute this response function, an SPGP uses a pseudo data set of size $M < N$, such that
\begin{equation}
    \mathbf{\mathrm{X_{SPGP}}} = \{\mathbf{\mathrm{X_{GP}}}\}_{m=1}^M,
\end{equation}
where $\mathbf{\mathrm{X_{SPGP}}}$ is the set of input data used to compute the prediction in an SPGP and $\mathbf{\mathrm{X_{GP}}}$ is the set of input data used to compute the predicted response in a standard GP. Hence, the spacing between pseudo data points is known. Moreover, $||\mathbf{\mathrm{X_{SPGP}}}||=M$ and $||\mathbf{\mathrm{X_{GP}}}||=N$. This reforged structure of a GP into an SPGP enables computing time decreases on the order of $O(N^3) \rightarrow O(NM^2)$ since $M < N$.

Another method to decrease the computing time of BO is efficient global optimization (EGO) \cite{Jones1998, Hennig2012, Jeong2005}. Similar to standard BO, EGO implements a surrogate model to generate the input-output response function, however, EGO can acquire a global optimum in fewer online iterations than BO by bounding the derivatives of the acquisition function relative to either the target variable or the surrogate standard error \cite{Jones1998}:
\begin{align}
\begin{split}
    \frac{\delta \mathrm{EI}(\mathbf{\mathrm{X}})}{\delta y(\mathbf{\mathrm{X}})} &< 0 \quad \mathrm{and} \\
    \frac{\delta \mathrm{EI}(\mathbf{\mathrm{X}})}{\delta s(\mathbf{\mathrm{X}})} &> 0,
\end{split}
\end{align}
where $\mathrm{EI}$ is the Expected Improvement acquisition function defined in the next section, $y$ is the target response variable and $s$ is the standard error of the surrogate. Additionally, van Stein \textit{et al.} \cite{van2019automatic} further decrease the computing time of EGO by parallelizing the computation of the gradients.

In order to decrease the computing time of BO, the methods mentioned above either (1) make significant changes to the surrogate model or (2) rely on computing the gradients of the AF to bound the search space. A downfall of computing the gradients of an AF is immediately constraining the AF of choice to be differentiable. Thus, the EGO studies above are constrained to using only the expected improvement AF and cannot use other AFs, even if it may be more advantageous. Therefore, in this study, we implement methods of decreasing the computing time of BO that do not constrain the user to select a certain surrogate model or AF to run the procedure. Instead, the method used in this paper is a lightweight implementation of a simple space-bounding method, not requiring gradients, from which new experiments are acquired. Furthermore, selective pruning of memory data from outside the bounded search space drives a significant decrease in BO compute time relative to standard BO. Hence, the method described in the next section supports the use of any surrogate model or AF. In this paper, we illustrate the computing times achieved using a GP surrogate model and four different AFs.

\section{Methods}

\begin{figure*}[h]
    \begin{center}
    \includegraphics[width=1.65\columnwidth]{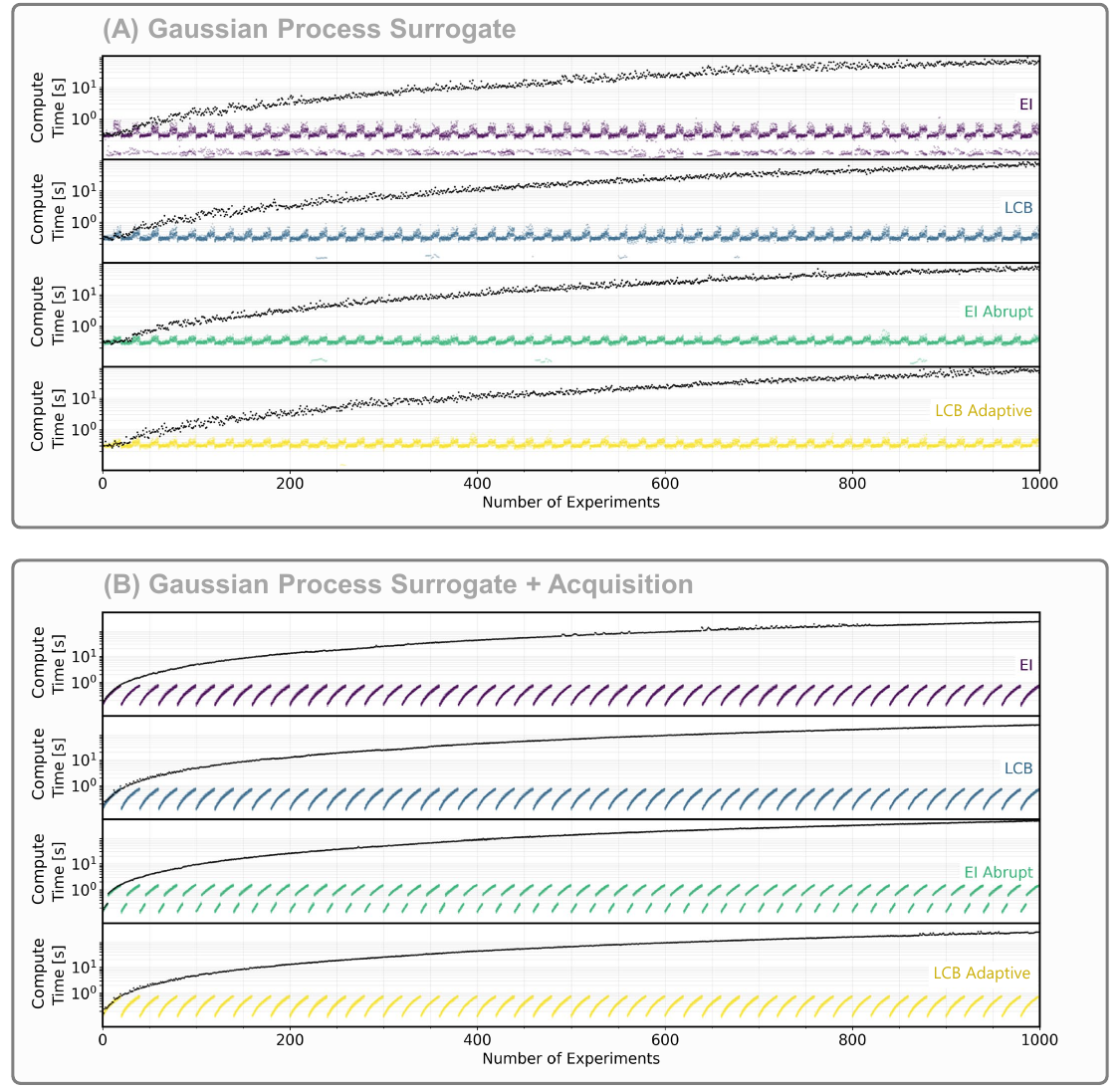}
    \end{center}
       \caption{Computing times of a Bayesian optimization procedure with a Gaussian process surrogate model. (A) Wall-clock computing times  per experiment for only the GP computation component across $N$ data points on a 6D analytical Ackley function. (B) Wall-clock computing times per experiment for the acquisition of new data from a GP computed across a mesh grid of 10k data points on a 6D analytical Ackley function. Each panel represents the computing times of four AFs, from top to bottom: EI, LCB, EI Abrupt, and LCB Adaptive. The colored scatter points represent the compute times per experiment of twelve independent optimization procedures using the memory pruning ZoMBI method for each AF. The black scatter points represent the benchmark compute times per experiment of one independent optimization using standard BO for each AF. For ZoMBI, $N\leq20$ \textit{via} memory pruning and for standard BO, $N$ is the number of experiments. All compute times are wall-clock compute times measured from the MIT Supercloud Nvidia Volta V100 GPUs. The $y$-axes are shown in log scale.}
    \label{fig:gp_compute}
    \end{figure*}

In this paper, we extend the implementation of a BO wrapper developed by Siemenn \textit{et al.} \cite{Siemenn2023} that bounds the acquisition and search space while pruning old memory data that lay outside of these computed bounds. This approach is entitled Zooming Memory-Based Initialization (ZoMBI) and is described further in \cite{Siemenn2023} with code publicly available. In brief, for a minimization objective, $f$, the bounds for each dimension, $d$, are computed uniquely based on the $\mathbf{\mathrm{min}}(\mathbf{\mathrm{X}}_d)$ and the $\mathbf{\mathrm{max}}(\mathbf{\mathrm{X}}_d)$ of the $m$ best-performing memory points, \textit{i.e.}, the points that achieve the $m$ lowest target $f$ values, from the set $\mathbf{\mathrm{X}}$. For every loop, all data points that lie outside of the constrained space will be pruned from memory. This is computationally favorable because as the search bounds iteratively zoom in, the target space inside the bounds increases in resolution by the surrogate model while all other space decreases in resolution. A standard GP surrogate model as well as a neural network (NN) surrogate model are used in the ZoMBI optimization procedure to demonstrate its generalizability to several unique surrogate models.

The computing times of four unique AFs implemented using ZoMBI are benchmarked against their standard BO counterparts. These four AFs are expected improvement (EI), lower confidence bound (LCB), EI Abrupt, and LCB Adaptive. Each of these mathematical figures of merit uniquely balances the exploitation of surrogate posterior means and the exploration of surrogate posterior variances.


EI is defined as \cite{Mockus1978, ei, Zhan2020}:
\begin{align}
\begin{split}
    \label{eq:ei}
    a_\textrm{EI}(\mathbf{\mathrm{X}}, \mathbf{\mathrm{Y}}; \xi, \eta)=& \left(\mu(\mathbf{\mathrm{X}}) - \mathbf{\mathrm{min}}(\mathbf{\mathrm{Y}}) - \xi\right)\Phi(Z)+\sigma(\mathbf{\mathrm{X}})\psi(Z),  \\
    \mathrm{where} \quad Z =& \frac{\mu(\mathbf{\mathrm{X}}) - \mathbf{\mathrm{min}}(\mathbf{\mathrm{Y}}) - \xi}{\sigma(\mathbf{\mathrm{X}})},
\end{split}
\end{align}
where $\mathbf{\mathrm{X}}$ is the set of input data $\{x_1, x_2,...x_N\}$, $x_j\in \mathbb{R}^d$ for $d$ dimensions, $\mathbf{\mathrm{Y}}$ is the set of corresponding response values $\{y_1, y_2,...y_N\}$, $y_j\in \mathbb{R}$, $\xi$ is a hyperparameter tuned to favor exploration or exploitation of the surrogate, and $\Phi(\cdot)$ and $\psi(\cdot)$ are the normal cumulative and probability density functions, respectively. EI strikes a balance between exploration and exploitation while considering the prior best-performing response variable of the set, $\mathbf{\mathrm{min}}(\mathbf{\mathrm{Y}})$.

LCB is defined as \cite{Auer2002, Cox1992}:
\begin{equation}
    \label{eq:lcb}
    a_\textrm{LCB}(\mathbf{\mathrm{X}}; \beta) = \mu(\mathbf{\mathrm{X}})-\beta{\sigma(\mathbf{\mathrm{X}})},
\end{equation}
where $\beta$ is a hyperparameter tuned to factor exploration or exploitation of the surrogate means, $\mu$, and variances $\sigma$. A higher $\beta$ favors exploration of surrogate variances while a lower $\beta$ favors exploitation of surrogate means.

EI Abrupt is defined as \cite{Siemenn2023}:
\begin{align}
\begin{split}
    \label{eq:abrupt}
    &a_\textrm{EI Abrupt}(\mathbf{\mathrm{X}}, \mathbf{\mathrm{Y}}; \beta, \xi, \eta)=\\
    \; &
    \begin{cases}
        a_\textrm{EI}(\mathbf{\mathrm{X}}, \mathbf{\mathrm{Y}}; \xi, \eta), & \text{if } |\Delta \{y_{N-3...N}\}|\leq \eta\\
        a_\textrm{LCB}(\mathbf{\mathrm{X}}; \beta), & \text{otherwise}
    \end{cases}
\end{split}
\end{align}
where the mode of acquisition is abruptly switched between EI and LCB depending on if the finite difference between the $\{y_{N-3...N}\}$ previous response values is below a hyperparameter threshold, $\eta$. EI Abrupt provides another level of tunable exploration-exploitation by actively swapping between these modes as more data is collected.

LCB Adaptive is defined as \cite{Siemenn2023, Srinivas2010, phoenics}:
\begin{equation}
    \label{eq:ada}
    a_\textrm{LCB Adaptive}(\mathbf{\mathrm{X}}, N; \beta, \epsilon) = \mu(\mathbf{\mathrm{X}})-\epsilon^N\beta{\sigma(\mathbf{\mathrm{X}})},
\end{equation}
where the hyperparameter, $\beta$, is actively tuned as the number of collected data points, $N = ||\mathbf{\mathrm{X}}||$, increases. LCB Adaptive exponentially decays from being more explorative to then becoming more exploitative as $N$ increases. Since ZoMBI actively prunes the set $\mathbf{\mathrm{X}}$, which decreases $N$ until more experiments are collected, LCB Adaptive is always switching acquisition modes throughout the optimization procedure.


In this paper, we demonstrate the computing times of each of these AFs implemented with the ZoMBI bounding and pruning method as well as implemented with just standard BO. The computing times are further bifurcated into (1) the surrogate model compute times per experiment alone on $N$ data points and (2) the surrogate + AF compute times per experiment on a mesh grid of 10k data points since acquisition of new data points requires the computation of the surrogate across a mesh grid of points in the space.

First, the computing times of each AF with a GP surrogate model are measured on a 6-dimensional analytical Ackley function \cite{ackley1987} for 1000 experiments for both ZoMBI and standard BO implementations. Second, the computing times of just the ZoMBI AF implementations with a NN surrogate model for 200 experiments are measured on a real-world 5-dimensional data set of inorganic crystalline material band gaps, available as open-access from Materials Project \cite{Jain2013}. Both of these experiments are run on the high-performance supercomputer, MIT Supercloud to measure the wall-clock computing times of both the surrogate models and the acquisition functions \cite{reuther2018interactive}.

\section{Results}


\begin{figure}[t]
    \begin{center}
    \includegraphics[width=0.8\columnwidth]{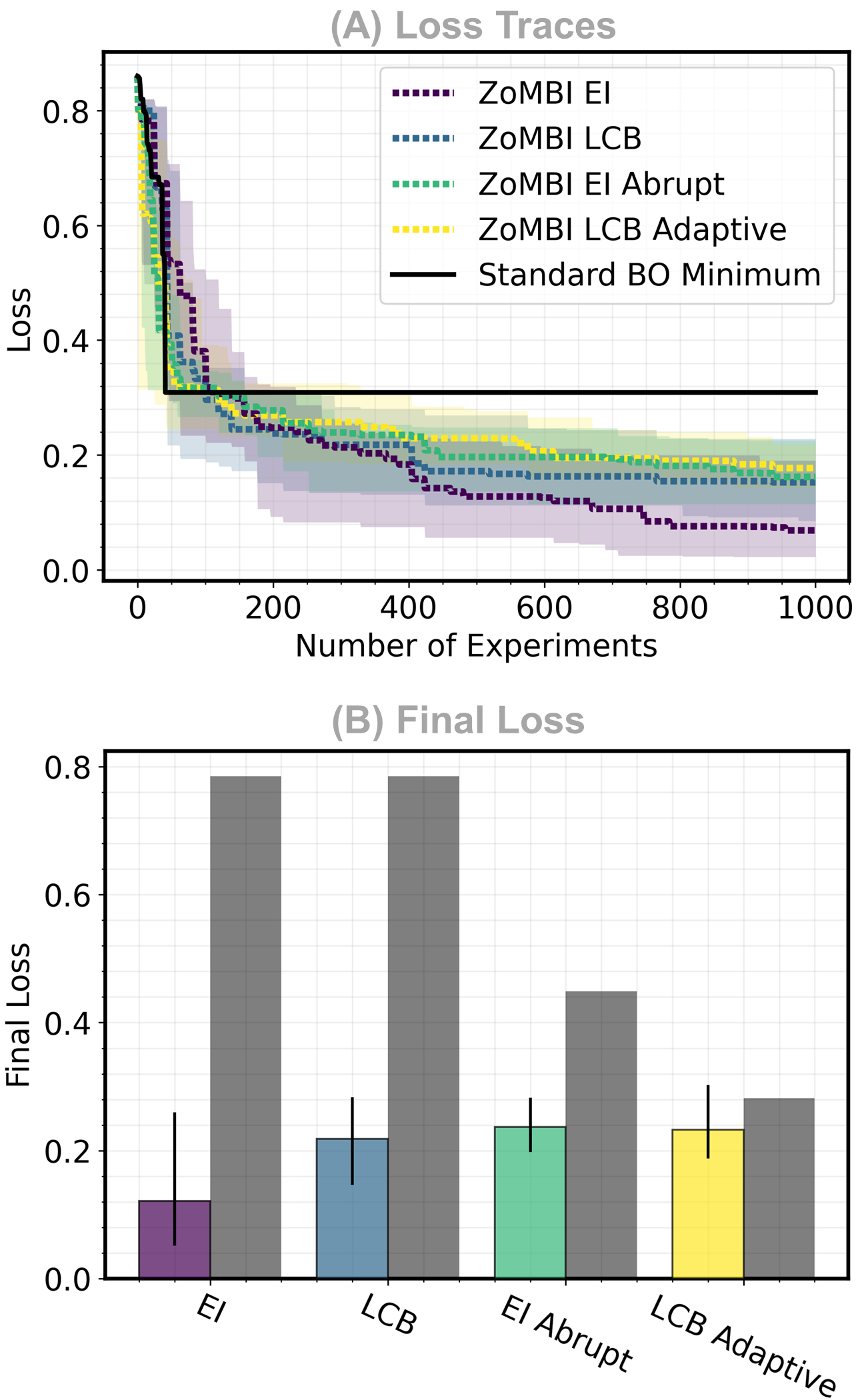}
    \end{center}
        \caption{Loss of standard and ZoMBI Bayesian optimization on a 6-dimensional Ackley function. (A) Loss traces over the 1000-experiment optimization procedure from Figure \ref{fig:gp_compute}. Only the minimum standard BO loss trace is shown for clarity. (B) Final loss values after 1000 sampled experiments. The colored bars and traces illustrate the median minimum values discovered by twelve independent trials of the memory pruning ZoMBI method for each AF with the 5th and 95th percentile indicated by (A) the shaded region and (B) the error bars. The grey bars illustrate the minimum values discovered by one independent trial of standard BO for each AF.}
    \label{fig:response}
    \end{figure}

\begin{figure*}[h]
    \begin{center}
    \includegraphics[width=1.65\columnwidth]{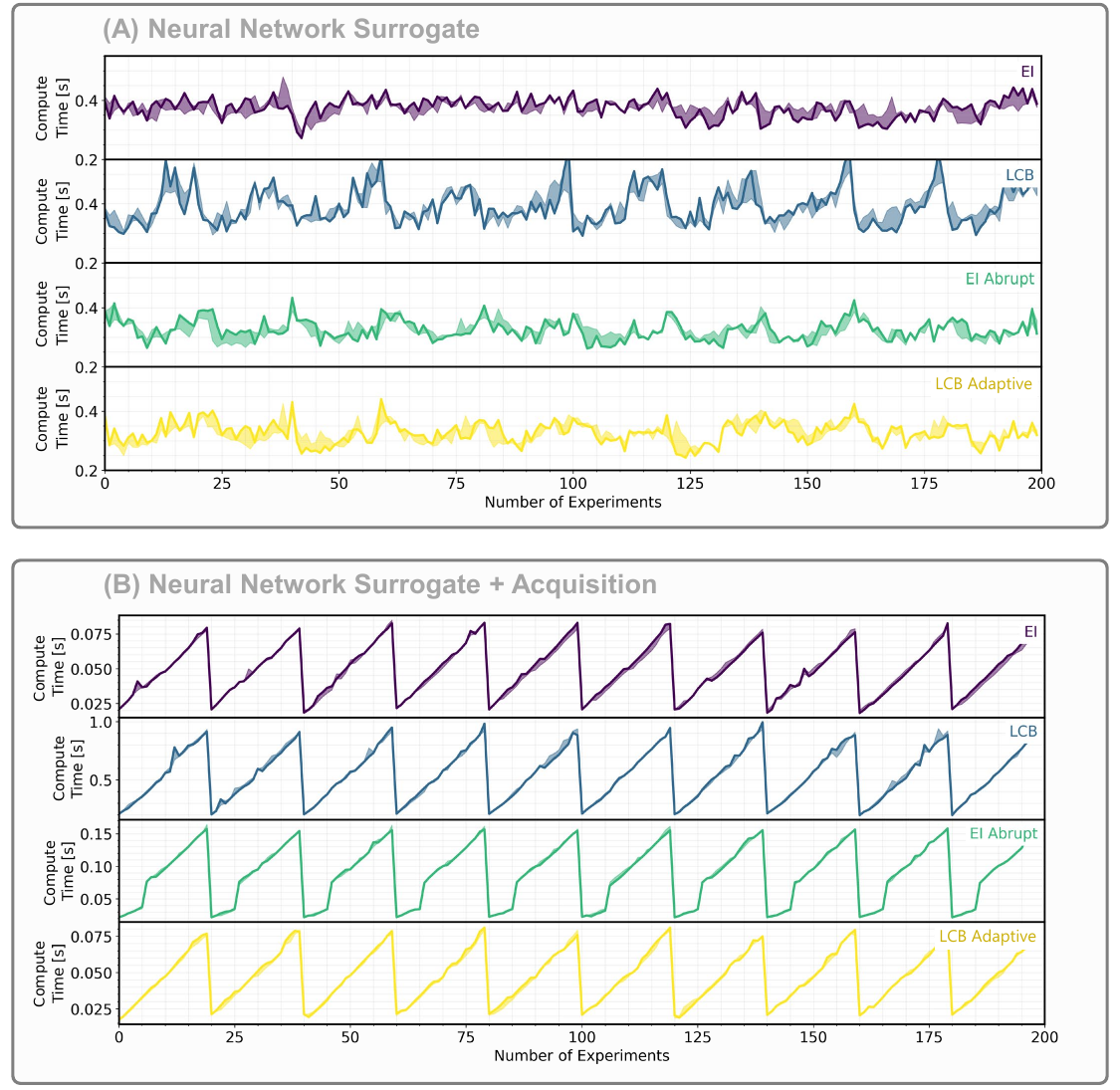}
    \end{center}
       \caption{Computing times of a Bayesian optimization procedure with a pre-trained neural network surrogate model. (A) Wall-clock computing times per experiment for only the NN computation component across $N$ data points on a 5D real-world materials data set. (B) Wall-clock computing times per experiment for the acquisition of new data from a NN computed across a mesh grid of 10k data points on a 5D real-world materials data set. Each panel represents the computing times of four AFs, from top to bottom: EI, LCB, EI Abrupt, and LCB Adaptive. The median values are shown by the solid line and the 5th and 95th percentile range is shown by the shaded region. All compute times are wall-clock compute times measured from the MIT Supercloud Nvidia Volta V100 GPUs.}
    \label{fig:nn_compute}
    \end{figure*}

\subsection{Gaussian Process Surrogate on a 6D Analytical Data Set}
In this section, we demonstrate a decrease in BO computing times, relative to standard BO, using the ZoMBI method of memory pruning on an \textit{in silico} optimization experiment of an analytical 6-dimensional Ackley function  \cite{Siemenn2023, ackley1987}. This \textit{in silico} optimization experiment is run on the MIT Supercloud Nvidia Volta V100 GPU\cite{reuther2018interactive}. 

Figure \ref{fig:gp_compute} illustrates (A) the time to compute a GP surrogate model for each iterative experiment and (B) the time to acquire new data points by computing the GP surrogate across a mesh grid for each iterative experiment. One operation is included in the measurement of GP compute time: (1) the fitting of training data, $\mathbf{\mathrm{X}}$, to a GP model using a mixture of kernel functions, in this case, Matern 5/2 kernels. Two operations are included in the measurement of GP + acquisition compute time: (1) the prediction and storage of the response values $\mathbf{\mathrm{Y}}$ from the GP for a mesh grid of 10k data points from a bounded set of $\mathbf{\mathrm{X}}$ and (2) the computation of the acquisition figure of merit from one of Equations \ref{eq:ei}--\ref{eq:ada}, hence, GP + acquisition computing times are higher than just GP computing times alone. 

In Figure \ref{fig:gp_compute}(A), a square wave pattern in ZoMBI computing times is shown by the colored scatter points. This is a result of the ZoMBI process selecting the top performing acquired experiments every $N=20$ experiments and pruning the rest from the memory to bound the surrogate mesh grid computation. Hence, for every 20 experiments, a drop in computing times is noted for ZoMBI, whereas computing times for standard BO continue to increase polynomially per experiment as additional experiments are collected to calculate the GP. As a result of this memory pruning, ZoMBI computing times per experiment demonstrate a non-increasing trend, even after 1000 experiments are acquired. Therefore, the memory pruning procedure significantly decreases the computing time per experiment relative to standard BO. Furthermore, a low spread between scatter points plotted from each of the twelve independent trials demonstrates the high reproducibility of results using the MIT Supercloud GPUs.


Similar to the GP surrogate computing time results, a significant decrease in computing times using memory pruning is demonstrated for the acquisition of new data points, shown in Figure \ref{fig:gp_compute}(B). A log-transformed sawtooth pattern is shown between memory pruning steps where a drop in compute time occurs. Again, low variance between the twelve independent trials is demonstrated due to the high overlap between scatter points. Using ZoMBI, the time to compute the GP surrogate alone is approximately 0.2 seconds per experiment (Figure \ref{fig:gp_compute}(A)), while the time to compute and acquire new data points from the surrogate model takes approximately 1 second per experiment (Figure \ref{fig:gp_compute}(B)). This difference arises due to the number of computations being performed: Figure \ref{fig:gp_compute}(A) computes the GP across only $N\leq20$ points while Figure \ref{fig:gp_compute}(B) computes the GP and the acquisition value (Equations \ref{eq:ei}--\ref{eq:ada}) across 10k points in a mesh grid to acquire new data points. After 1000 experiments are collected, the ZoMBI method still achieves computing times of 1 second per experiment, however, the standard BO method polynomially approaches compute times of 100 seconds per experiment, a factor of 100x slower. Therefore, computing times of BO are significantly reduced using a memory pruning and bounded optimization approach. But, does this pruning and bounding process adversely impact optimization performance?



Figure \ref{fig:response} illustrates the convergence of ZoMBI and standard BO on the global minimum of the 6-dimensional Ackley function. Not only does this memory pruning and bounded optimization procedure not adversely impact optimization performance, but it is also demonstrated to outperform standard BO on the 6-dimensional Ackley function. ZoMBI EI achieves the lowest function values after 1000 experiments with LCB, then EI Abrupt, then LCB Adaptive following, in that order. The reverse is noted for standard BO. This implies that without the memory pruning and search space bounding features of ZoMBI, the actively adapting acquisition functions, EI Abrupt and LCB Adaptive, perform better than the conventional EI and LCB acquisition functions. Moreover, we note that standard BO, shown as the black trace in Figure \ref{fig:response}(A), stops learning after fewer than 50 experiments due to local minima and the sharpness of the Ackley function global minimum \cite{ackley1987, merkuryeva2011benchmark}
while all ZoMBI methods continue to learn by continuously zooming in the search space bounds.

\subsection{Neural Network Surrogate on a 5D Real-world Data Set}

In this section, we demonstrate a decrease in BO computing times, relative to standard BO, using the ZoMBI method of memory pruning on an optimization problem translatable to \textit{in situ} experimentation. The data set optimized is a 5-dimensional open-access data set of inorganic crystals with the objective of optimizing the properties density, formation energy, energy above hull, Fermi energy to find a material with 1.4eV band gap \cite{Jain2013, DeJong2015}. A pre-trained NN is used as the surrogate model instead of the GP to demonstrate the generalizability of the memory pruning method to various surrogate models.

Figure \ref{fig:nn_compute} illustrates the time to fit a pre-trained NN to the set $\mathbf{\mathrm{X}}$ on the MIT Supercloud \cite{reuther2018interactive}. Figure \ref{fig:nn_compute}(A) illustrates fitting the NN surrogate to a maximum of $N=20$ points using ZoMBI, whereas Figure \ref{fig:nn_compute}(B) illustrates fitting the NN and computing the respective AF to a mesh grid of 10k points using ZoMBI. The combination of the NN fitting to few data and also being pre-trained produces a noisy trace of computing times in Figure \ref{fig:nn_compute}(A). However, as the number of fitting points increases from 20 to 10k, a much clearer trend in computing times can be seen in Figure \ref{fig:nn_compute}(B). 


Similar to the GP surrogate results on the 6D Ackley function in Figure \ref{fig:gp_compute}, a sawtooth pattern, resetting every 20 experiments is shown for the NN + AF computing times in Figure \ref{fig:nn_compute} for the 5D real-world data set. Although each of these AFs has a similar structure to their compute time curves, each $y$-axis has a different scale, and LCB is noted to have the highest computing time. This is likely due to LCB's explorative nature constantly generating a wide search bound which encompasses many more data points when compared to any of the greedier AF methods. Furthermore, an interesting pattern is seen in the EI Abrupt curves where the first rising segment has a different structure than the second rising segment, this is the abrupt switch between EI and LCB sampling modes that changes the bounding and, in turn, changes the number of data points kept in memory. 

Overall, the NN surrogate run on the 5D real-world data set produces similar non-increasing computing times per experiment to the GP surrogate run on the 6D analytical Ackley function. Hence, demonstrating the potential for the ZoMBI memory pruning and bounding optimization method to be generalizable to various surrogate models, without modification, to decrease the computing time of BO.

\section{Summary \& Conclusions}

In this paper, we demonstrate the capabilities of search space bounding and memory pruning in Bayesian optimization to significantly decrease the optimization procedure's computing time. We demonstrate this decrease in compute time by up to 100x across two unique data sets, two unique surrogate models, and four unique acquisition functions, all of which are run on the high-performance MIT Supercloud supercomputer \cite{reuther2018interactive}. 

The method of bounding and memory pruning using Zooming Memory-Based Initialization (ZoMBI) \cite{Siemenn2023} implemented in this paper takes the best-performing memory points and uses those values to construct a constrained search region for the acquisition function to sample from. Upon consecutive constraints, prior data points that lay outside of these bounds are pruned from memory, decreasing the number of data points used to fit a surrogate model, in turn, decreasing the time required to compute the surrogate model and its acquisition function.

We demonstrate that this iterative constraining and pruning process achieves a sawtooth computing time pattern per experiment, relative to standard BO that exhibits a polynomially increasing computing time trend following $O(N^3)$ for $N$ experiments. The sawtooth computing time pattern is shown to reset back to near-zero after each memory pruning update, hence, producing a non-increasing computing time trend per experiment. Furthermore, this decreased computing time is shown to persist across analytical and real-world data sets, across Gaussian Process regression and neural network surrogate models, and across four acquisition functions: expected improvement, lower confidence bound, abrupt expected improvement, and adaptive lower confidence bound. The results demonstrated in this paper are also shown to be reproducible with low variance across several independent trials by being run on the MIT Supercloud supercomputer. Hence, in this paper, we demonstrate the reproducibility and generalizability of the proposed ZoMBI memory pruning and bounded optimization method to decrease the computing times of Bayesian optimization across a variety of data sets, surrogate models, and acquisition functions.

\bibliographystyle{IEEEtran}
\bibliography{references.bib}

\end{document}